\documentclass[conference]{IEEEtran}
\IEEEoverridecommandlockouts
\usepackage{cite}
\usepackage{amsmath,amssymb}
\usepackage{graphicx}
\usepackage{booktabs}
\usepackage{balance}
\usepackage{url}
\usepackage[hidelinks]{hyperref}

\begin{document}

\title{Auditing Quality Filters for Long-Tail Human Data Curation}

\author{\IEEEauthorblockN{Rishav Agarwal}
\IEEEauthorblockA{\textit{Gritt Robotics} \\
rishav@gritt.ai}
\and
\IEEEauthorblockN{Nirshal Chandra Sekar}
\IEEEauthorblockA{\textit{Gritt Robotics} \\
nsekar@gritt.ai}
\and
\IEEEauthorblockN{Anirudh Vemula}
\IEEEauthorblockA{\textit{Gritt Robotics} \\
anirudh@gritt.ai}}

\maketitle

\begin{abstract}
Robots on construction sites must detect workers who are kneeling or bending, which we call \emph{low poses}. These workers can be lost from training datasets during automatic labeling. We study a pipeline that detects people, estimates their body joints using NLF~\cite{nlf}, and groups similar poses. \emph{Low poses} account for only about $2\%$ of the retained examples. This low share may partly reflect the pipeline’s quality filter, which rejects examples with low detection confidence or uncertain joint estimates. We examine this filtering using four alternative pose clues: bounding-box shape, vertical body span, pose grouping aligned to the scene’s vertical direction, and image appearance. All four suggest that \emph{low poses} are rejected by the filter more often. Separately, controlled simulated scenes show that a person detector fine-tuned on a public construction dataset misses more workers in these poses even when we correct their bounding box height is matched to that of standing workers. These findings suggest that \emph{low poses} are scarce and hard to find, and we cannot rely on bounding boxes or poses for long-tail human data curation.

\end{abstract}

\begin{IEEEkeywords}
long-term perception, human pose, construction robotics, dataset shift, rare classes
\end{IEEEkeywords}

\section{Introduction}
\label{sec:intro}

\begin{figure}[t]
\centering
\includegraphics[width=\columnwidth]{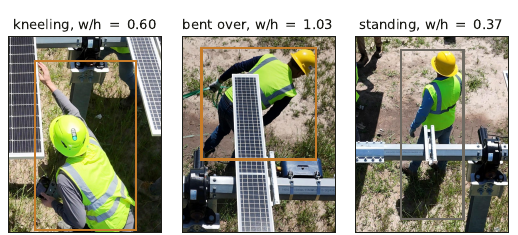}
\caption{\emph{Low poses} in the mined library, with the box the detector
produced. A worker kneeling at a tracker motor and a worker bent over a panel
(orange), next to a standing worker for comparison (grey). The width-to-height
ratio above each panel is the quantity the box-shape check in
Section~\ref{sec:why} uses.}
\label{fig:lowpose}
\end{figure}

Gritt robots operate on active solar-construction sites, moving among crews who guide panels and work beneath the solar array. Reliable person detection must cover difficult cases, including workers kneeling behind material piles or partially hidden by structural supports. We use \emph{low poses} to describe kneeling, squatting, and deeply bent-over body configurations (Fig.~\ref{fig:lowpose}). The term describes posture, not visibility or a measured distance above the ground: a worker in a \emph{low pose} may be fully visible or partially occluded. These poses form part of the dataset’s \emph{long tail}, with far fewer examples than common standing poses.

These poses are rare in the curated dataset, so a natural response is to collect and annotate more examples. However, collecting more footage may not address how examples are selected. Manually labeling body joints across months of site video is expensive, so automated pipelines estimate them instead~\cite{nlf,moon2022neuralannot} and often discard uncertain estimates. ~\emph{Low poses} can be difficult to reconstruct, especially when they appear small or are partly hidden. A filter intended to improve pose-label reliability may therefore discard examples with usable person boxes. The resulting dataset mixes two sources of scarcity: how often a pose appears in the footage and how often it survives filtering. We call this the \emph{selection--scarcity confound}.

We investigate this problem through two analyses. First, we examine people already found by the detector and ask whether the pose quality filter rejects \emph{low poses} more often than other poses. Second, we render the same scenes with people in different poses and measure how often a detector fine-tuned on a public construction dataset finds them. The first analysis examines which examples survive annotation; the second examines which poses are better detected.

Finally, since the pipeline’s own pose labels may be unreliable for checking what the filter discarded. We therefore compare four alternative visual and geometric clues for identifying \emph{low poses}, and examine how the pipeline’s body-alignment step can also fail. In simulation, we separately test changes in character appearance, pose, and the height of a person’s bounding box in pixels. Together, these analyses motivate retaining valid person boxes for detection training even when the corresponding pose estimates are uncertain.

\section{Pose Mining and Observed Scarcity}
\label{sec:tail}

We process the images with a person detector and a pose estimator. Candidate person boxes come from the person detector bundled with the Neural Localizer Fields multi-person model (\texttt{nlf\_l\_multi} 0.3.2)~\cite{nlf}, which also estimates body-joint positions and their uncertainty using the SMPL representation~\cite{smpl}. The detector proposes $7K$ candidate people in a $4.3K$ subset of a  $17K$ image dataset. We retain estimates with detector confidence of at least $0.5$ and mean joint uncertainty of at most $150$\,mm, which keeps $3.9K$ images.

\begin{figure}[t]
\centering
\includegraphics[width=\columnwidth]{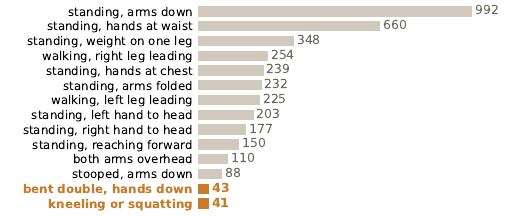}
\caption{The fourteen pose groups mined from $3{,}762$ detections. The two
low-pose groups contain 41 and 43 instances.}
\label{fig:tail}
\end{figure}

The estimated joints are expressed in camera coordinates, so camera tilt can make similar poses appear differently oriented. To correct for this, we estimate the vertical direction in each image from the median ankle-to-pelvis axis of detected people and align the bodies to it. We then represent each pose by the coordinates of 20 joints relative to the body root, producing a 60-dimensional vector. We normalize this vector by body extent and rotate it to a common hip orientation so that grouping depends on body arrangement rather than position, scale, or facing direction. Finally, we remove near-repeated poses in neighboring frames, dropping a body whose vector L2 norm lies within $0.08$ of one already kept in the preceding $90$ frames, which gives us $3.7K$ images. We cluster the remaining vectors into fourteen groups, each named after a representative example. The kneeling/squatting and bent-double groups define our \emph{low poses} and together account for about $2\%$ of the retained library (Fig.~\ref{fig:tail}).

\begin{table}[t] \caption{Low-pose share by capture source.} \label{tab:source} \centering \footnotesize \setlength{\tabcolsep}{3pt} \begin{tabular}{lrrl} \toprule Capture source & Inst. & Low & Share [95\% CI] \\ \midrule Gritt dataset & 3{,}258 & \phantom{0}31 & \phantom{0}0.95\% [0.54, \phantom{0}1.56] \\ External construction dataset & \phantom{3,}504 & \phantom{0}53 & 10.52\% [7.24, 14.20] \\ \midrule Full retained library & 3{,}762 & \phantom{0}84 & \phantom{0}2.23\% [1.51, \phantom{0}3.23] \\ \bottomrule \end{tabular}

\vspace{2pt}
{\footnotesize The apparent frequency of the tail depends on where
the images were collected. Intervals are 95\% bootstrap intervals clustered by
capture recording.\par}
\end{table}

\paragraph*{Where images come from affects pose coverage}

Low poses account for a larger share of retained examples in external construction images than in  camera footage from our deployments (Table~\ref{tab:source}). These sources differ in camera placement, the work being performed, and which images were selected. The comparison shows that pose coverage varies with the image source, but does not identify the cause or establish how common \emph{low poses} are across construction sites.

As neighboring video frames often show the same workers under similar conditions, we compute confidence intervals by resampling whole recordings rather than individual detections. External still images are treated as separate units. The filter audit includes only people already found by the detector and it cannot account for people the detector missed.

\section{Auditing Pose-Dependent Filtering}
\label{sec:why}

Most quality-filter rejections are due to uncertain joint estimates. To check whether \emph{low poses}    are rejected more often, we first assign each rejected pose estimate to its nearest pose group. This suggests that rejected examples contain a larger share of \emph{low poses}. However, the same uncertain joints determine both rejection and group assignment. Errors in those joints could make other poses appear to be \emph{low poses}. Adding comparable noise to trusted poses produced a similar share of low-pose labels, so the observed grouping does not establish that the filter preferentially rejects low poses.

\paragraph*{Alternative measurements}
We check the filter using four alternative clues for identifying \emph{low poses} (Table~\ref{tab:proxy}, rows 2--5; row 1 is the original pose labels). Each avoids some, but not all, of the weaknesses of the original pose labels:
\begin{itemize}

\item \textbf{Box shape:} The width-to-height ratio of the person’s box.

\item \textbf{Vertical body span:} The estimated skeleton’s vertical span, normalized by body extent, identifies folded bodies without assigning them to pose clusters.

\item \textbf{Scene-based alignment:} GeoCalib~\cite{geocalib} estimates vertical from the surrounding scene rather than from people’s bodies. We use this direction to align the estimated joints before assigning pose groups.

\item \textbf{Image appearance:} A classifier using DINOv2 features~\cite{dinov2} identifies \emph{low poses} from cropped person images rather than estimated joints.
\end{itemize}

All four checks suggest that the filter rejects \emph{low poses} more often than other poses. However, these checks can also misidentify poses, especially when people are difficult to see. Although the image classifier performs well on held-out recordings, we have not verified its accuracy on rejected examples. Manually checking poses in both accepted and rejected examples would help confirm the finding.

\paragraph*{Factors associated with filtering}

The filter rejects people whose boxes are wide relative to their height more often. This remains true after accounting for box height, contact with the image boundary, and image source. However, box shape can reflect either a low pose or partial occlusion, so pose alone may not explain the difference.

Allowing less certain joint estimates reduces this rejection gap. Changing the required detection confidence has much less effect. Among people already detected, the joint-uncertainty requirement therefore appears to drive much of the unequal filtering.

Together, these findings suggest that filtering makes \emph{low poses} look rarer in the retained dataset. Their frequency before filtering remains uncertain because the pose checks can mislabel examples, and removing repeated poses also changes the counts.

\begin{table}[t] \caption{Filter pass rate for \emph{low poses}, measured five ways.} \label{tab:proxy} \centering \footnotesize \setlength{\tabcolsep}{3.5pt} \begin{tabular}{llc} \toprule Check & Estimated joints used & Ratio \\ \midrule Original pose labels & pose vector and alignment & \textit{(0.32)} \\ Box shape & none & 0.26 \\ Vertical body span & span only & 0.43--0.47 \\ Scene-based alignment & pose vector only & 0.66 \\ Image appearance & none & 0.50 \\ \bottomrule \end{tabular}

\vspace{2pt}
{\footnotesize Filter pass rate for examples identified as \emph{low poses},
divided by the pass rate for the remaining examples. A ratio of $1$ indicates
equal pass rates. The first row is excluded from our evidence because it is
sensitive to joint-estimation errors; row 2 uses untruncated boxes.\par}
\end{table}

\begin{figure}[tb]
\centering
\includegraphics[width=\columnwidth]{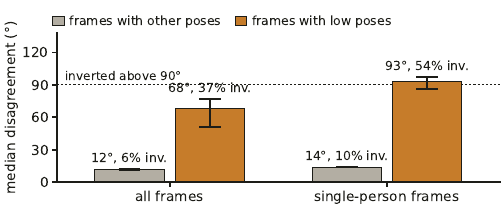}
\caption{The body-based estimate of vertical compared with a scene-based
gravity estimate~\cite{geocalib}. Bootstrap $95\%$ intervals.}
\label{fig:gravity}
\end{figure}

\paragraph*{Estimating vertical from people can fail}

Before grouping poses, the pipeline corrects for camera tilt by assuming that the direction from a person’s ankles to their pelvis points upward. This works best for standing people and can fail when someone kneels or bends over.

We compare this direction with GeoCalib’s estimate, which uses the surrounding scene. The two estimates disagree much more in frames containing \emph{low poses}, sometimes pointing in nearly opposite directions (Fig.~\ref{fig:gravity}). This also happens in frames with only one person.

An incorrect estimate of vertical can rotate a body incorrectly and place it in the wrong pose group. However, neither GeoCalib nor our pose labels provide ground truth, so this comparison reveals disagreement rather than measuring the exact alignment error. Using depth sensors to estimate vertical from the ground plane is an alternative we have not yet tested.

\section{Testing the Detector in Simulation}
\label{sec:sim}

\paragraph*{Experimental design}

\begin{figure}[t]
\centering
\includegraphics[width=\columnwidth]{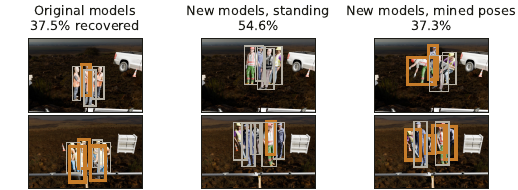}
\caption{Two scenes under three rendering conditions:
(a) original human models,
(b) new models standing upright, and
(c) the same new models in mined poses.
Ground-truth boxes are orange where the detector misses a person, and the
percentages are computed over the $968$ people present in all five conditions.
Changing the models changes recall for this detector, so the pose comparisons
hold the models fixed; changing their poses reduces recall and also changes
their image size.}
\label{fig:threeway}
\end{figure}

We now ask a separate question: how well does a person detector find people in the mined poses?
We render 200 construction scenes three times: first with the original human model, then with four new models standing upright, and finally with the new models in poses from the mined library (Fig.~\ref{fig:threeway}). Person locations, cameras, lighting, and the surrounding scene remain fixed. Comparing the first two versions tests whether changing the human models affects detection. Comparing the last two tests the effect of changing pose, which can also change bounding-box height and how much of the person is visible.

The detector is YOLO11s~\cite{yolo11}, pretrained on COCO~\cite{coco2014} and
fine-tuned for this paper on the public \texttt{rbyz/ppe-6-classes-ntld9}
dataset~\cite{ppe6classes}, using that export's own split and its person class
only: $200$ training images with $536$ person boxes, and $50$ validation images
with $128$.
The validation images are used only to select the operating confidence
threshold, as the value maximizing $F_1$ on them ($0.45$, at which recall on
those images is $87.5\%$).
It has seen no synthetic imagery and none of our own data, so these renders are
entirely out of sample.
We measure \emph{recall}, the fraction of rendered people it finds, over $968$ person
instances matched across all five rendering conditions.
A prediction counts as a detection when its box overlaps the ground-truth
person box with intersection-over-union of at least $0.5$.

\paragraph*{Pose changes also change bounding-box height}
Replacing the original human models with the new ones, still standing, raises recall by about 17 percentage points: the assets themselves matter to this detector, so the later comparisons hold them fixed. Applying the mined poses to those same models then reduces recall by about 17 points.

However, a crouching person usually occupies fewer vertical pixels than
a standing person at the same location.
The initial comparison therefore measures the effect of changing pose
together with changes in bounding-box height, body outline, and visibility.

To test whether smaller bounding boxes explain the detector’s difficulty, we render two additional conditions: posed figures enlarged to match the box height of standing figures, and standing figures shrunk to match the box height of posed figures.

We restrict the comparison to 130 people across 79 scenes. In all five rendering conditions, each selected person’s box must overlap every other person’s box by at most IoU $0.3$ and remain at least $4$\,px from every image edge. This reduces the effects of crowding and image-boundary clipping.

Within this subset, matching bounding-box height reduces but does not eliminate the recall gap (Table~\ref{tab:render}). It falls from about $29$ points at the original scale to about $18$ points when standing figures are shrunk to the posed size, and to about $25$ points when posed figures are enlarged to the standing size; both intervals exclude zero. We compute $95\%$ bootstrap confidence intervals by resampling whole scenes, keeping people from the same scene together. Scoring the same detector at full render resolution rather than its training resolution leaves the shrunk-standing comparison unchanged ($-23.8$ pp, $[-32.5, -15.5]$). Matching box height does not match silhouette or visibility. For this detector, smaller box height alone therefore cannot explain the poorer detection of mined poses.
A zero-shot open-vocabulary detector, OWLv2~\cite{owlv2} prompted with
\emph{person} and \emph{worker} and thresholded on the same validation images,
points the same way over all $968$ people ($-14.7$ pp for the pose change and
$-10.0$ pp at matched height), although on the restricted subset its intervals
include zero.

\begin{table}[t]
\caption{Changes in recall between rendering conditions.}
\label{tab:render}
\centering\footnotesize
\setlength{\tabcolsep}{3pt}
\begin{tabular}{lrr}
\toprule
Comparison & Change (pp) & 95\% interval \\
\midrule
Original to new upright models & +17.1 & [+13.6, +20.6] \\
Standing to mined poses & -17.4 & [-20.3, -14.4] \\
\midrule
Restricted subset: original scale & -29.2 & [-38.1, -20.8] \\
Restricted subset: standing shrunk & -17.7 & [-25.6, -10.2] \\
Restricted subset: posed enlarged & -24.6 & [-33.3, -16.4] \\
\bottomrule
\end{tabular}

\vspace{2pt}
{\footnotesize Changes are in percentage points (pp);
negative values indicate lower recall.
The last three rows use the restricted subset and should be compared
with each other rather than with the full-scene result. The two matched-height
rows scale in opposite directions, which changes mutual occlusion in opposite
ways; the gap survives both.}
\end{table}

\paragraph*{Implications}
Bounding-box height alone does not explain either result.
Matching it leaves most of the detector's recall gap in simulation, and
accounting for it leaves the field filter's association with box shape largely
unchanged.
The detector and the pose-quality filter therefore require separate checks:
the same postures are both harder to detect and less likely to survive
annotation.

Simulated scenes let us test specific detection failures and could provide additional training examples. We have not tested whether training on these examples improves detection in real construction footage. The paired pose comparisons keep the human models and surrounding scene fixed, but changing pose can also change body outline and visibility. The results show that this detector performs worse on mined poses, even when bounding-box height is matched.

\section{Related Work}
\label{sec:related}

\paragraph*{Rare-case mining in autonomous driving}
Mining difficult cases and improving their coverage are established goals.
CODA provides a road corner-case detection benchmark~\cite{coda2022}.
AIDE combines issue discovery, data curation, automatic labeling, model updating, and verification for autonomous-driving detection~\cite{aide2024}.
Our study examines whether a pose-annotation filter selectively removes a within-person pose group, and whether the labels used to audit that removal are themselves reliable. 

\paragraph*{Selection bias and latent subgroups}
Class-balanced self-training addresses dominance of large classes during pseudo-label generation~\cite{cbst2018}, while the DataComp audit demonstrates that quality filtering can change group representation~\cite{hong2024}.
Our contribution is a construction-robotics case study of this broader problem, including a noise control that undermines a naive audit and alternative measurements that corroborate selective rejection.

Hwang et al.\ study rare human poses using clustering and targeted learning strategies~\cite{hwang2020}.
Hidden stratification explains how aggregate accuracy can conceal subgroup failures~\cite{oakdenrayner}, and GEORGE uses learned features to discover latent subclasses~\cite{sohoni}.
These motivate pose-level analysis, but recovering a pose group from model outputs does not by itself validate its frequency among rejected annotations.

\paragraph*{Synthetic evaluation and application context}
Synthetic human datasets support perception training~\cite{surreal,motsynth,bedlam}.
Counterfactual simulation testing~\cite{ruiz2022}, Virtual KITTI~\cite{vkitti}, and 3DB~\cite{threedb} establish rendering as a tool for controlled model analysis.
We apply that approach to construction poses and bounding-box height.
Off-road pedestrian detection~\cite{pezzementi2018} and construction datasets~\cite{soda} provide the closest application context.
Our focus is the reliability of pose labels used to audit rejected examples in a construction-robotics pipeline.

\section{Implications and Limitations}
\label{sec:disc}

\paragraph*{Keep valid boxes even when pose estimates fail}
A body can have a usable person box but an unreliable 3D reconstruction.
Whether an example is kept for person-detection training should depend on the validity of its box, not on whether its pose estimate passes the quality filter.
Pose labels can retain a stricter criterion.
This does not mean accepting every detector prediction: false positives and localization errors still require quality control.
The proposed separation remains to be evaluated end to end.

\paragraph*{Audit what is discarded}
Report which population each frequency describes: raw images, detected candidates, quality-filtered examples, or deduplicated poses.
Validate proxy labels on a sample from both sides of the filter and report pose group admission alongside label quality.
Relaxing an uncertainty threshold trades coverage against fidelity; the present measurements do not identify a universally appropriate operating point.

\paragraph*{Evaluate important pose groups}

Alongside recall, report how often the detector finds people in each important pose group, with the number of examples and confidence intervals. Breaking results down by bounding-box height, how much of the person is hidden, and image source can help explain when detection fails. Also, report precision at the chosen confidence threshold. Reporting precision by pose requires a rule for assigning false detections to pose groups, which we do not investigate here. These measures assess detector performance but do not establish the coverage of the  robot~\cite{salay,mltestscore}.

\paragraph*{What remains unresolved}

Our filter audit includes only people already found by the detector. It uses alternative pose checks whose accuracy on rejected examples has not been independently verified. We also test only one pose estimator, and GeoCalib’s estimate of vertical may be wrong. In simulation, resizing people can change how much they hide one another, which is why we match size in both directions and report both, and our restricted comparison may not represent crowded scenes. The next steps are to manually label accepted and rejected examples, repeat the audit with another pose estimator, and test on real images not used during training.

The practical lesson is that keeping only reliable pose estimates can remove useful examples for person detection. We therefore need to check both which examples the filter discards and which people the detector misses, with results reported separately for important pose groups.

\balance
\bibliographystyle{IEEEtran}
\bibliography{refs}

@inproceedings{coda2022,
  author={Li, Kaican and Chen, Kai and Wang, Haoyu and Hong, Lanqing and
          Ye, Chaoqiang and Han, Jianhua and Chen, Yukuai and Zhang, Wei and
          Xu, Chunjing and Yeung, Dit-Yan and Liang, Xiaodan and Li, Zhenguo and Xu, Hang},
  title={{CODA}: A Real-World Road Corner Case Dataset for Object Detection in Autonomous Driving},
  booktitle={European Conference on Computer Vision (ECCV)},
  year={2022},
  note={arXiv:2203.07724}
}

@inproceedings{aide2024,
  author={Liang, Mingfu and Su, Jong-Chyi and Schulter, Samuel and Garg, Sparsh
          and Zhao, Shiyu and Wu, Ying and Chandraker, Manmohan},
  title={{AIDE}: An Automatic Data Engine for Object Detection in Autonomous Driving},
  booktitle={IEEE/CVF Conference on Computer Vision and Pattern Recognition (CVPR)},
  year={2024},
  note={arXiv:2403.17373}
}

@inproceedings{cbst2018,
  author={Zou, Yang and Yu, Zhiding and Kumar, B. V. K. Vijaya and Wang, Jinsong},
  title={Unsupervised Domain Adaptation for Semantic Segmentation via Class-Balanced Self-Training},
  booktitle={European Conference on Computer Vision (ECCV)},
  year={2018},
  note={arXiv:1810.07911}
}

@article{pezzementi2018,
  author  = {Pezzementi, Zachary and Tabor, Trenton and Hu, Peiyun and
             Chang, Jonathan K. and Ramanan, Deva and Wellington, Carl and
             Babu, Benzun P. Wisely and Herman, Herman},
  title   = {Comparing Apples and Oranges: Off-Road Pedestrian Detection on the
             {NREC} Agricultural Person-Detection Dataset},
  journal = {Journal of Field Robotics},
  volume  = {35},
  number  = {4},
  year    = {2018},
  note    = {arXiv:1707.07169}
}

@article{salay,
  author  = {Salay, Rick and Czarnecki, Krzysztof and Kuwajima, Hiroshi and
             Yasuoka, Hirotoshi and Nakae, Toshihiro and Abdelzad, Vahdat and
             Huang, Chengjie and Kahn, Maximilian and Nguyen, Van Duong},
  title   = {The Missing Link: Developing a Safety Case for Perception
             Components in Automated Driving},
  journal = {SAE International Journal of Advances and Current Practices in
             Mobility},
  volume  = {5},
  number  = {2},
  pages   = {567--579},
  year    = {2022},
  note    = {arXiv:2108.13294}
}

@inproceedings{mltestscore,
  author    = {Breck, Eric and Cai, Shanqing and Nielsen, Eric and
               Salib, Michael and Sculley, D.},
  title     = {The {ML} Test Score: A Rubric for {ML} Production Readiness and
               Technical Debt Reduction},
  booktitle = {IEEE International Conference on Big Data},
  year      = {2017}
}

@article{smpl,
  title={{SMPL}: A Skinned Multi-Person Linear Model},
  author={Loper, Matthew and Mahmood, Naureen and Romero, Javier and
          Pons-Moll, Gerard and Black, Michael J.},
  journal={ACM Transactions on Graphics},
  volume={34}, number={6}, pages={248:1--248:16}, year={2015}
}

@inproceedings{nlf,
  title={Neural Localizer Fields for Continuous {3D} Human Pose and Shape Estimation},
  author={S{\'a}r{\'a}ndi, Istv{\'a}n and Pons-Moll, Gerard},
  booktitle={Advances in Neural Information Processing Systems (NeurIPS)},
  year={2024}
}

@inproceedings{geocalib,
  title     = {{GeoCalib}: Learning Single-image Calibration with Geometric Optimization},
  author    = {Veicht, Alexander and Sarlin, Paul-Edouard and
               Lindenberger, Philipp and Pollefeys, Marc},
  booktitle = {European Conference on Computer Vision (ECCV)},
  year      = {2024}
}

@inproceedings{surreal,
  title     = {Learning from Synthetic Humans},
  author    = {Varol, G{\"u}l and Romero, Javier and Martin, Xavier and Mahmood, Naureen and Black, Michael J. and Laptev, Ivan and Schmid, Cordelia},
  booktitle = {IEEE Conference on Computer Vision and Pattern Recognition (CVPR)},
  year      = {2017}
}

@inproceedings{motsynth,
  title     = {{MOTSynth}: How Can Synthetic Data Help Pedestrian Detection and Tracking?},
  author    = {Fabbri, Matteo and Bra{\v{s}}o, Guillem and Maugeri, Gianluca and Cetintas, Orcun and Gasparini, Riccardo and O{\v{s}}ep, Aljo{\v{s}}a and Calderara, Simone and Leal-Taix{\'e}, Laura and Cucchiara, Rita},
  booktitle = {IEEE/CVF International Conference on Computer Vision (ICCV)},
  year      = {2021}
}

@inproceedings{moon2022neuralannot,
  title={Neuralannot: Neural annotator for 3d human mesh training sets},
  author={Moon, Gyeongsik and Choi, Hongsuk and Lee, Kyoung Mu},
  booktitle={2022 IEEE/CVF Conference on Computer Vision and Pattern Recognition Workshops (CVPRW)},
  pages={2298--2306},
  year={2022},
  organization={IEEE}
}

@inproceedings{bedlam,
  title     = {{BEDLAM}: A Synthetic Dataset of Bodies Exhibiting Detailed Lifelike Animated Motion},
  author    = {Black, Michael J. and Patel, Priyanka and Tesch, Joachim and Yang, Jinlong},
  booktitle = {IEEE/CVF Conference on Computer Vision and Pattern Recognition (CVPR)},
  year      = {2023}
}

@article{soda,
  title   = {{SODA}: A Large-scale Open Site Object Detection Dataset for Deep Learning in Construction},
  author  = {Duan, Rui and Deng, Hui and Tian, Mao and Deng, Yichuan and Lin, Jiarui},
  journal = {Automation in Construction},
  volume  = {142},
  pages   = {104499},
  year    = {2022}
}

@inproceedings{vkitti,
  author    = {Gaidon, Adrien and Wang, Qiao and Cabon, Yohann and Vig, Eleonora},
  title     = {Virtual Worlds as Proxy for Multi-Object Tracking Analysis},
  booktitle = {IEEE Conference on Computer Vision and Pattern Recognition (CVPR)},
  year      = {2016}
}

@inproceedings{threedb,
  author    = {Leclerc, Guillaume and Salman, Hadi and Ilyas, Andrew and
               Vemprala, Sai and Engstrom, Logan and Vineet, Vibhav and
               Xiao, Kai and Zhang, Pengchuan and Santurkar, Shibani and
               Yang, Greg and Kapoor, Ashish and Madry, Aleksander},
  title     = {{3DB}: A Framework for Debugging Computer Vision Models},
  booktitle = {Advances in Neural Information Processing Systems (NeurIPS)},
  year      = {2022}
}

@inproceedings{hong2024,
  author    = {Rachel Hong and William Agnew and Tadayoshi Kohno and
               Jamie Morgenstern},
  title     = {Who's in and who's out? A case study of multimodal
               {CLIP}-filtering in {DataComp}},
  booktitle = {ACM Conference on Equity and Access in Algorithms, Mechanisms,
               and Optimization (EAAMO)},
  year      = {2024},
  note      = {arXiv:2405.08209}
}

@article{hwang2020,
  author  = {Jihye Hwang and John Yang and Nojun Kwak},
  title   = {Exploring rare pose in human pose estimation},
  journal = {IEEE Access},
  volume  = {8},
  pages   = {194964--194977},
  year    = {2020}
}

@inproceedings{oakdenrayner,
  author    = {Luke Oakden-Rayner and Jared Dunnmon and Gustavo Carneiro and Christopher R\'{e}},
  title     = {Hidden stratification causes clinically meaningful failures in machine
               learning for medical imaging},
  booktitle = {ACM Conference on Health, Inference, and Learning (CHIL)},
  year      = {2020}
}

@inproceedings{sohoni,
  author    = {Nimit S. Sohoni and Jared A. Dunnmon and Geoffrey Angus and Albert Gu and
               Christopher R\'{e}},
  title     = {No subclass left behind: Fine-grained robustness in coarse-grained
               classification problems},
  booktitle = {Advances in Neural Information Processing Systems (NeurIPS)},
  year      = {2020}
}

@inproceedings{ruiz2022,
  author    = {Nataniel Ruiz and Sarah Adel Bargal and Cihang Xie and Kate Saenko and
               Stan Sclaroff},
  title     = {Finding differences between transformers and {ConvNets} using
               counterfactual simulation testing},
  booktitle = {Advances in Neural Information Processing Systems (NeurIPS)},
  year      = {2022}
}

@article{dinov2,
  author  = {Maxime Oquab and Timoth\'{e}e Darcet and Th\'{e}o Moutakanni and
             Huy V. Vo and Marc Szafraniec and others},
  title   = {{DINOv2}: Learning robust visual features without supervision},
  journal = {Transactions on Machine Learning Research},
  year    = {2024},
  note    = {arXiv:2304.07193}
}

@misc{yolo11,
  author={Jocher, Glenn and Qiu, Jing},
  title={Ultralytics {YOLO11}},
  year={2024},
  howpublished={\url{https://github.com/ultralytics/ultralytics}}
}

@inproceedings{coco2014,
  author={Lin, Tsung-Yi and Maire, Michael and Belongie, Serge and Hays, James and
          Perona, Pietro and Ramanan, Deva and Doll\'{a}r, Piotr and Zitnick, C. Lawrence},
  title={{Microsoft COCO}: Common Objects in Context},
  booktitle={European Conference on Computer Vision (ECCV)},
  year={2014}
}

@misc{ppe6classes,
  author={{rbyz}},
  title={{PPE} 6 Classes Dataset},
  howpublished={Roboflow Universe, \url{https://universe.roboflow.com/rbyz/ppe-6-classes-ntld9}},
  note={Accessed 2026-09-06}
}

@inproceedings{owlv2,
  author={Minderer, Matthias and Gritsenko, Alexey and Houlsby, Neil},
  title={Scaling Open-Vocabulary Object Detection},
  booktitle={Advances in Neural Information Processing Systems (NeurIPS)},
  year={2023}
}

\end{document}